\documentclass{styles/svproc}

\usepackage[a4paper,asymmetric]{geometry}

\usepackage{listings}
\usepackage{hyperref}

\usepackage{wrapfig}

\usepackage{algorithm}
\usepackage{algpseudocode}
\usepackage{xcolor}
\usepackage{tcolorbox}
\tcbuselibrary{minted, skins, breakable}
\usepackage{minted}
\usepackage{mathtools}
\usepackage[utf8]{inputenc}
\usepackage[T1]{fontenc}
\usepackage[english]{babel}
\usepackage{booktabs}
\usepackage{amsfonts}
\usepackage{nicefrac}
\usepackage{microtype}
\usepackage{amsmath,  amssymb}
\usepackage{fancybox}
\usepackage{graphicx}
\usepackage{float}
\usepackage{color}
\usepackage{tikz}
\usepackage{array}
\usepackage{subcaption}

\usepackage{enumitem}
\usepackage{wrapfig}
\usepackage{enumitem}
\usepackage{etoolbox}
\usepackage{diagbox}
\usepackage{comment}

\usepackage{makecell}
\usepackage{multirow}
\usepackage{bbm}
\usepackage{mathpazo}

\usepackage[framemethod=TikZ]{mdframed}
\mdfsetup{skipabove=5pt,
innertopmargin=0pt}

\newcommand{\todo}[1]{{\color{red} #1}}

\newcommand{\shortskip}{\vspace{3pt}}

\newcommand{\ie}{i.e.\ }
\newcommand{\eg}{e.g.\ }

\newcommand{\delete}[1]{}

\begin{document}
\mainmatter
\title{Actively Resolving Contextual Uncertainty \\for Underspecified  Tasks in Natural Language}
\titlerunning{Actively Resolving Contextual Uncertainty }
\author{Zachary Ravichandran\inst{1},
Jonathan Diller\inst{1},
Fernando Cladera\inst{1}, \\
Varun Murali\inst{2},
George J. Pappas\inst{1}, \and
Vijay Kumar\inst{1}
}
\authorrunning{Ravichandran et al.}

\tocauthor{Zachary Ravichandran}

\institute{
GRASP Laboratory, University of Pennsylvania
\and
Electrical \& Computer Engineering, Texas A\&M University
\\
Correspondence to \email{zacravi@seas.upenn.edu}
}

\maketitle

\begin{abstract}
Foundation models provide robots with the ability to interpret natural language and reason about environmental context, yet most language-conditioned policies assume that goals are well-specified and that task-relevant information is provided upfront via a prior map.
Operating in unfamiliar environments with underspecified tasks entails high contextual uncertainty---the robot must jointly infer what constitutes task success, what constitutes relevant information, and where (or whether) that information exists.
We address these limitations via \texttt{CLUE} (Closed-Loop contextual Uncertainty rEsolution), a framework for actively resolving contextual uncertainty given underspecified tasks in natural language.
\texttt{CLUE} uses an LLM-derived policy to hypothesize task-relevant concepts and potential plans. It then uses a language-embedded map, which is constructed online, to ground these hypotheses into actions.
The policy sequentially evaluates hypotheses via closed-loop environment interaction and refines its plans as it gathers new information.  
We deploy \texttt{CLUE}  on a Boston Dynamics Spot across three real indoor and outdoor environments spanning 15 tasks that require object disambiguation, functional inference, and occlusion reasoning.
\texttt{CLUE} achieves a success rate within 7 percentage points of an oracle policy and outperforms an LLM-enabled planner without closed-loop feedback by a 4x margin.
Supporting experiments demonstrate that simply building and then querying a language-enriched map is insufficient to resolve complex contextual planning tasks; these approaches achieve roughly one third the success rate of \texttt{CLUE} while requiring over 10x more VLM tokens. We provide additional information at \href{https://zacravichandran.github.io/CLUE}{https://zacravichandran.github.io/CLUE}

\end{abstract}

\section{Introduction}
\label{sec:intro}

\begin{figure}[t]
    \centering
    \includegraphics[width=1.00\linewidth]{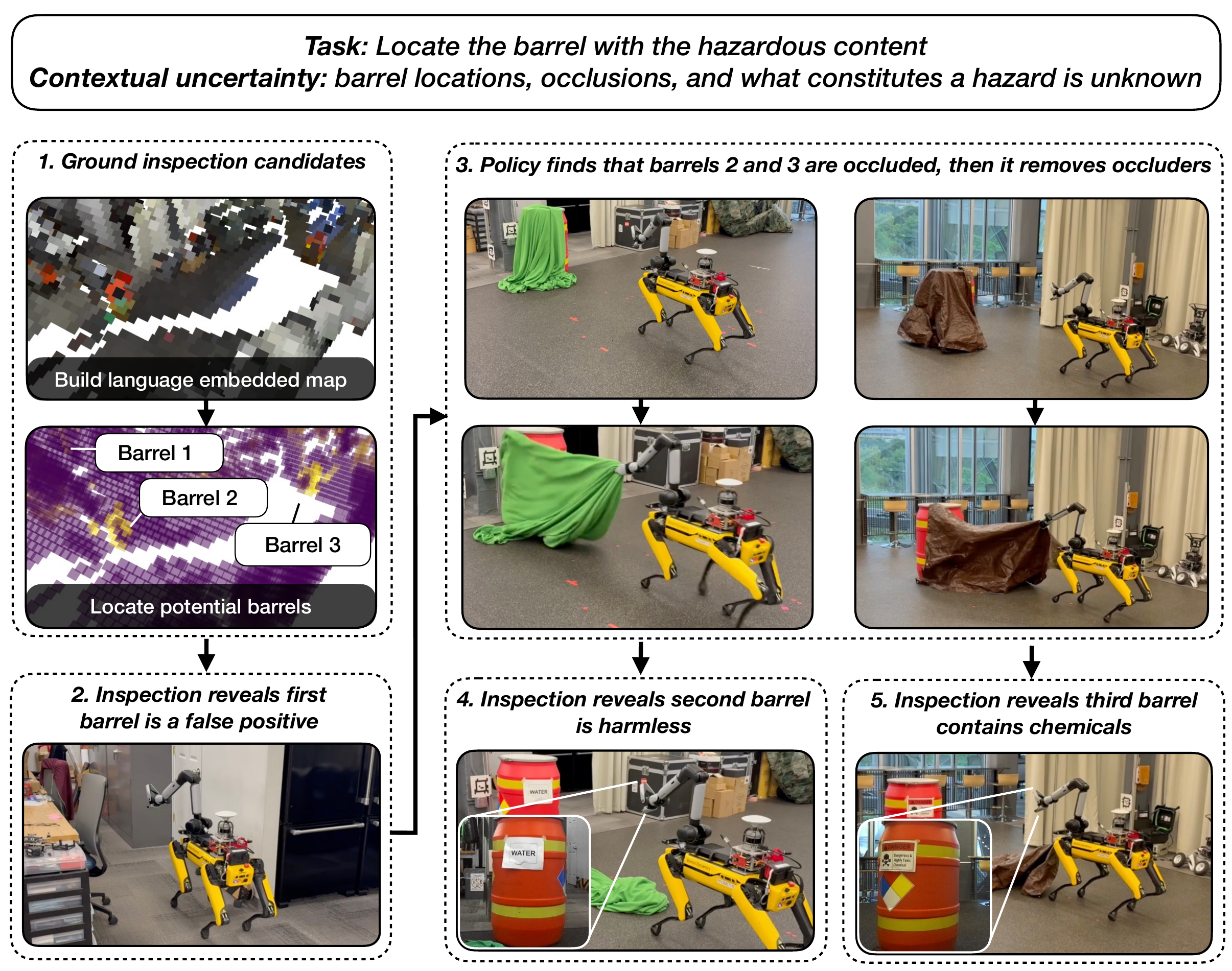}
    \caption{\textbf{Example Task.} The robot must identify a hazardous barrel, but it does not know barrel locations or what constitutes a hazard \emph{a priori}.
    It uses a language-embedded map constructed from RGB-D data to identify possible barrel locations. It rules out one as a false positive upon inspection. The other two candidates are barrels but are occluded by tarps. The robot removes the tarps, reads the signs hidden underneath, and finds the third barrel contains dangerous chemicals.}
    \label{fig:intro_figure}
    \vspace{-12pt}
\end{figure}

Consider a robot placed in an unfamiliar area and tasked to ``locate the barrel with the hazardous content,'' as illustrated in Figure~\ref{fig:intro_figure}.
Resolving this specification requires addressing several coupled forms of uncertainty about semantic (what constitutes a hazard, what constitutes success), metric (where are the relevant entities), and functional (occlusions prevent inspection but can be removed) information, which we refer to as \emph{contextual uncertainty}.
A successful robot will infer subtasks and relevant entities, navigate to inspect them, and resolve unexpected challenges---such as occlusions---that arise along the way.
While a large body of work builds language-aware planners using large language models (LLMs) and vision-language models (VLMs), these planners largely assume explicit task instructions and high-quality prior information, limiting their applicability in contextually uncertain scenarios~\cite{2026vaptamppaxton,wang2025vision}.

Classically, actively resolving uncertainty in task planning has been posed from a geometric perspective: the robot builds a metric map by selecting viewpoints that maintain localization accuracy or coverage~\cite{tao2025rt,zhou2021fuel}.
Such formulations reason about where the robot has and has not looked, but they are largely agnostic to contextual associations---that a tarp may conceal a barrel or labels reveal hazard levels.
Conversely, current language-driven planning work largely treats map retrieval as sufficient for task grounding.
Open-vocabulary map representations let a robot query a pre-built map~\cite{chen2023open} or online mapper~\cite{yokoyama2024vlfm,ong2025atlas} with free-form language, but they assume a single query returns the correct answer and offer no mechanisms to verify detections.
Dense scene-description methods caption every object and region encountered during mapping~\cite{Gorlo2025DAAAM}, but because captioning happens without knowledge of the task, the resulting descriptions are generic, expensive to produce, and still cannot prompt the robot to \emph{act}---\eg move a tarp when it is occluding task-relevant information.
A parallel line of work casts language-driven planning under uncertainty as belief-space or POMDP planning with foundation models~\cite{sun2023interactiveplanningusinglarge,zhao2025seeing,tang2025trupomdp,llm_pomdp_corl,kim2026large,he2026legs}.
These methods are largely  designed for information retrieval or manipulation tasks in constrained workspaces, and they do not demonstrate the closed-loop integration of hypothesis refinement, targeted inspection, and physical intervention required for real-world contextually uncertain tasks.
While they reason over uncertainty in object existence or location, they assume the task-relevant semantics are known \emph{a priori}.
In contextually uncertain tasks, relevant semantics are themselves latent variables that can only be resolved through closed-loop interaction with the environment.

We address these limitations via \texttt{CLUE} (Closed-Loop contextual Uncertainty rEsolution), a closed-loop framework for resolving contextual uncertainty that couples a language-embedded map with an LLM-enabled policy that iteratively replans as it gathers information.
Given an underspecified instruction, \texttt{CLUE} uses an LLM to hypothesize contextually relevant information and potential actions.
A language-embedded map is used to ground these hypotheses into concrete information-gathering goals; if a hypothesis is too precise to be grounded in the map, the LLM may broaden its hypothesis (e.g., ``red backpack'' to ``backpack'' to ``bag''). 
\texttt{CLUE} then actively validates these hypotheses through closed-loop interaction with the environment via VLM inspection, navigation, and manipulation. 
\texttt{CLUE} uses observations acquired from interaction to update its hypothesis---a tractable approximation to belief for open-vocabulary environments that captures its understanding of the state (confirmed entities, what remains uncertain, and what must be verified) and goal conditions. 
As \texttt{CLUE} gathers new information, its hypotheses evolve to drive plan generation. 
To summarize, our contributions are:
\begin{enumerate}
    \item A closed-loop framework  that couples LLM-enabled contextual inference and language-embedded map grounding, enabling active information gathering for  resolving contextual uncertainty. 
     \item A grounding mechanism that translates LLM-generated hypotheses into concrete planning objectives via a language-embedded map.
    \item Feedback mechanisms via language-embedded map querying and targeted VLM inspection to resolve hypotheses and gather task-specific information. 
\end{enumerate}

We experimentally validate these contributions through fifteen tasks requiring \texttt{CLUE} to disambiguate objects based on contextual relevance, reason over functional affordances, and resolve occlusions observed online, all performed on a Boston Dynamics Spot robot in three real-world environments.
We find that \texttt{CLUE} achieves a task success rate of within 7 percentage points of an oracle and over four times that of an LLM-enabled policy without closed-loop feedback.
In supporting experiments, \texttt{CLUE} achieves over three times the success rate of a state-of-the-art scene understanding framework which builds a language-enriched map \emph{a priori} then queries it for information---demonstrating that closed-loop environment interaction is necessary for resolving contextual uncertainty.

\section{Related work}
\label{sec:related_work}
We now overview three key areas of related work.

\shortskip
\noindent \textbf{LLM-Driven Planning.}
A primary methodological divide in works that leverage foundation models for planning is whether the model directly generates control sequences or plans over abstraction (\ie behaviors) that are realized by downstream controllers. For single-robot systems, recent Vision-Language-Action (VLA) models favor direct end-to-end control, as demonstrated in~\cite{Zitkovich2023RT,black2025pi0,li2025llara}, where the language agent maps open-world task descriptions and visual inputs directly to low-level motor commands, a trend mirrored in other single-agent~\cite{Ravichandran2025SPINE,chen2023open} and multi-agent~\cite{Kannan2024SMART,Chen2024Scalable} planners.
While direct LLM planning excels at implicit semantic reasoning, it often lacks formal safety guarantees and struggles with high-frequency reactive constraints. In contrast, works such as~\cite{Mandi2024RoCo,exploreeqa2024,Wu2025Hierarchical,ravichandran2025heterogeneous,dai2024optimal} position the LLM as a high-level translator that feeds objective weights, constraints, or 3D waypoints into robust classical solvers (e.g., MPC or RRT), ensuring safe, continuous execution while maintaining the LLM's semantic adaptability.

\shortskip
\noindent \textbf{Language-Driven Planning with Active Perception.}
Recent works have increasingly leveraged both LLMs and VLMs to handle active perception requirements during task planning in open-world environments. Several approaches model the problem in PDDL and utilize foundation models to resolve visual ambiguities or verify preconditions backward from a goal state during multi-step household tasks \cite{2026vaptamppaxton,Liu2026Natural,wang2025vision}. While these systems are able to ground language in visual observations, they generally require target objects to be in the robot's immediate field of view.
Other active perception methods focus on constructing and querying hierarchical scene graphs or other related semantic reconstructions of the environment~\cite{Gorlo2025DAAAM,Ravichandran2025SPINE,chen2023open}. These methods rely on a near-complete understanding of the environment prior to execution---assuming that map retrieval is sufficient to solve the task---and lack the ability to reason over partial observability.

\shortskip
\noindent \textbf{Language-Driven Planning with Uncertainty.}
A line of research extends LLM planning methods to incorporate notions of uncertainty~\cite{knowno2023,liangintrospective}.
A common approach is to 
frame robotic planning within belief spaces, a concept rooted in early discrete symbolic planning~\cite{srivastava2007abstraction}. These works often formulate the problem as a Partially Observable Markov Decision Process (POMDP). Recent frameworks utilize LLMs to generate POMDP transition and observation models, enabling robots to locate hidden objects~\cite{llm_pomdp_corl,tang2025trupomdp,he2026legs,kim2026large,sun2023interactiveplanningusinglarge}. Furthermore, multi-agent systems and LLM-based planners have been deployed to handle missing state variables during long-horizon manipulation tasks~\cite{nayak2024mapthor,zhao2025seeing,arnob2026effective}. 
Although these methods can handle various forms of uncertainty, they lack the cognitive flexibility to dynamically interpret the task-relevance of an object (e.g., determining that a barrel is a ``hazard'') or assess the affordances of an object as they update their semantic understanding of the scene (e.g., ascertaining that a cart can support a laptop). Unlike existing methods, our approach bridges this gap by reasoning over semantically embedded maps to locate conceptually relevant items, and subsequently interpreting objects and evaluating their physical appropriateness.

\section{Problem Definition}
\label{sec:problem}
 
A robot is placed in an unknown environment and must achieve an underspecified
task $\ell$ in natural language.
The robot is equipped with a set of high-level behaviors for navigation,
inspection, and manipulation, an odometry system, and an RGB-D camera.
We assume (i) the robot's pose is available from a reliable odometry
source, (ii) the environment is static over the course of an episode,
(iii) the information needed to satisfy $\ell$ is observable from RGB-D imagery
once the appropriate location is inspected, possibly after manipulation to
resolve occlusion, and (iv) the robot is given no predefined object catalog or
scene graph.
 
\shortskip
\noindent\textbf{POMDP formulation.}
We model this problem as a POMDP.
The state space $\mathcal{S}$ comprises the robot's pose and its surrounding
environment, including semantics.
The current state $s_i \in \mathcal{S}$ is unknown and must be estimated via
observations $o_i \in \mathcal{O}$, which are constructed from RGB-D
imagery upon executing a behavior.
The action space $\mathcal{A}$ encompasses robot behaviors
instantiated with referents in the environment.
Referents are properties of the environment, so only the subset
$\mathcal{A}(s_i) \subseteq \mathcal{A}$ is feasible in state $s_i$:
\eg \texttt{inspect}(\emph{barrel}) can be executed only if a barrel exists and is
reachable.
Because $s_i$ is not observed, the feasible set is itself uncertain: at each
iteration the policy proposes actions it \emph{believes} feasible, and a
grounding step rejects those that cannot be realized against the robot's current
environment representation (\S\ref{sec:info_extraction}).
Each task induces a latent set of goal states
$\mathcal{S}^\ell \subseteq \mathcal{S}$, \ie what constitutes task success is not
given explicitly but must be inferred from the instruction and observations
gathered online.
The robot must both reach a goal state and report task completion, which it does
via a terminal \texttt{answer}$(r)$ action whose argument $r$
conveys the task-required information:
\begin{equation}
\label{eq:reward}
	R(s_i, a_i) =
	\begin{cases}
		+1, & \text{if } a_i = \texttt{answer}(r) \text{ and } s_i \in \mathcal{S}^{\ell} \\
		-1, & \text{if } a_i = \texttt{answer}(r) \text{ and } s_i \notin \mathcal{S}^{\ell} \\
		\phantom{+}0, & \text{otherwise.}
	\end{cases}
\end{equation}
The episode terminates if \texttt{answer} is executed, or the action budget $N$ is exhausted.
 
\shortskip
\noindent\textbf{Objective.}
Planning under uncertainty may be addressed by maintaining a belief state
$b_i \in \mathcal{B}$, where $b_i(s_i) = \mathbb{P}(s_i \mid o_{1:i}, a_{1:i-1})$
is a probability distribution over the true state.
The robot chooses actions based on its policy $\pi \colon \mathcal{B} \to \mathcal{A}$,
and we seek
\begin{equation}
\label{eq:pi_opt}
    \pi^* = \arg\max_{\pi} \; \mathbb{E}
    \left[ \sum_{i=0}^{N-1} \gamma^i R(s_i, a_i) \;\middle|\; b_i, \pi \right]
\end{equation}
where $\gamma \in (0,1)$.
Because at most one reward is non-zero per episode,
Equation~\ref{eq:pi_opt} amounts to maximizing the probability of terminating in
a goal state while discouraging both superfluous information gathering and
premature answers.
 
\shortskip
\noindent\textbf{Contextual uncertainty.}
The underspecified task and unknown environment together induce
\emph{contextual uncertainty}, which manifests in three coupled dimensions.
\emph{Semantic uncertainty} concerns which entities are task-relevant and what
constitutes task success.
\emph{Metric uncertainty} concerns where those entities are located, and hence
which actions are feasible.
\emph{Functional uncertainty} concerns which physical interactions are required
to make task progress, \eg whether an occluding object must be moved before a
referent can be inspected.
Resolving contextual uncertainty amounts to maintaining a belief jointly over
$\mathcal{S}^\ell$, $s_i$, and consequently $\mathcal{A}(s_i)$, updated as
observations arrive.
 
\shortskip
\noindent\textbf{Tractable approximation.}
Maintaining this belief exactly is intractable: the state space is
continuous and open-vocabulary, and folding the unknown goal conditions
$\mathcal{S}^\ell$ into the state only compounds its dimensionality.
We therefore approximate the belief via a \emph{hypothesis} state
$h_i \in \mathcal{H}$, a symbolic account of the task, discovered entities, and
outstanding uncertainties, whose structure is detailed in \S\ref{sec:planner}
and whose fields correspond directly to the three dimensions above.
This yields a hypothesis update
\begin{align}
    \tau \colon (h_{i-1}, o_i, \ell) \mapsto h_i,
\end{align}
paired with a policy over hypotheses
\begin{align}
    \pi \colon h_i \mapsto a_i,
\end{align}
both of which \texttt{CLUE} realizes with a single query to a
pre-trained LLM (\S\ref{sec:planner}).
The entities carried in $h_i$ are precisely those over which the policy proposes
behaviors, and grounding against the environment representation determines which
of those proposals are feasible.
This does not solve Equation~\ref{eq:pi_opt} exactly; rather, it serves as a
tractable approximation that replans as each new observation reduces the robot's
uncertainty about the current state $s_i$ and goal conditions $\mathcal{S}^\ell$.

\section{Actively Resolving Contextual Uncertainty with CLUE}
\label{sec:framework}

\begin{figure}[!t]
    \centering
    \includegraphics[width=1.00\linewidth]{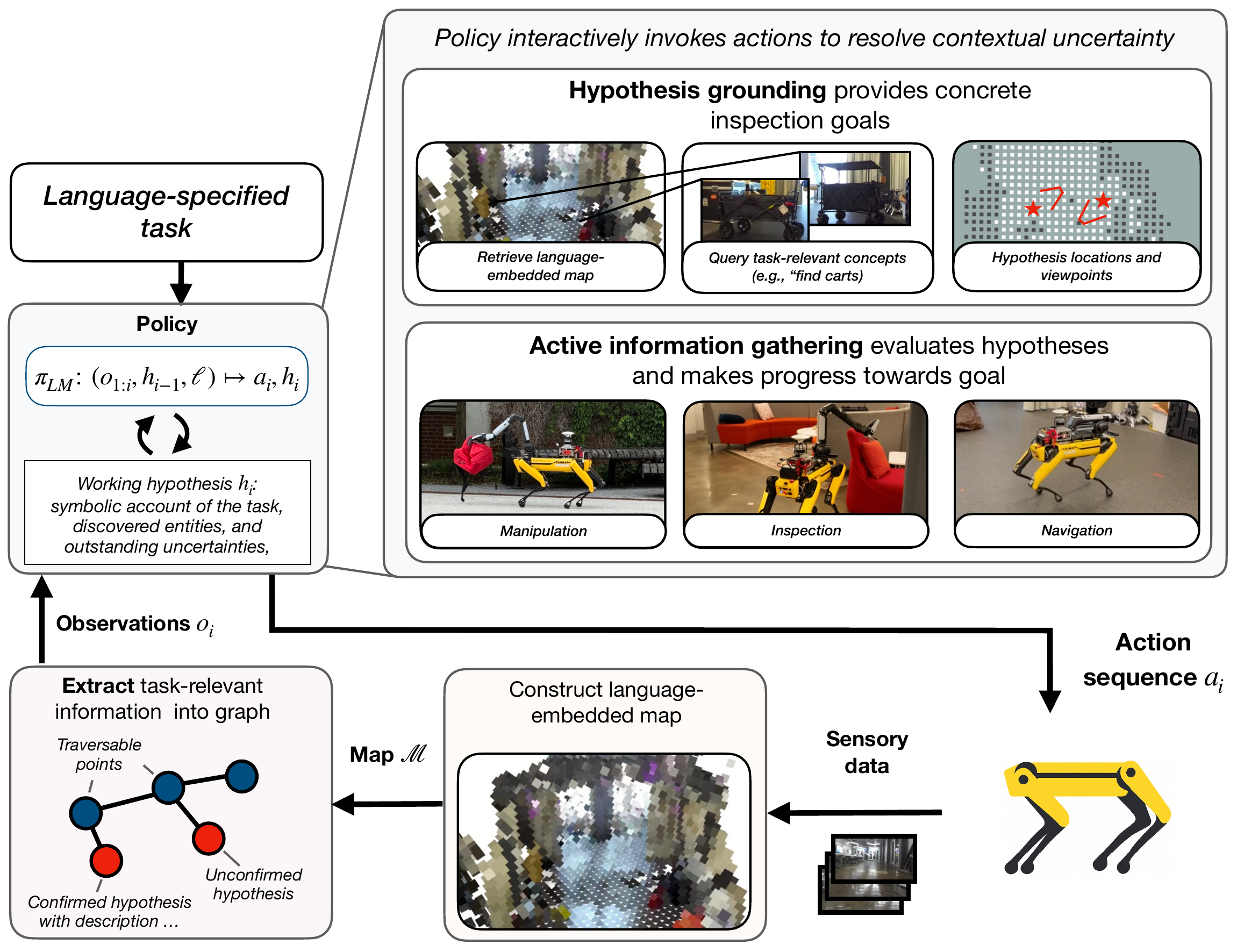}
    \caption{\textbf{CLUE architecture.} \texttt{CLUE} uses an LLM as a policy to realize underspecified language tasks in a closed-loop manner via two action types---hypothesis grounding and active information gathering. The LLM then constructs actions comprising manipulation, inspection, and navigation to resolve contextual uncertainties and make task progress. Information from the physical world is provided to the LLM via a symbolic graph-based observation space.}
    \label{fig:method}
    \vspace{-12pt}
\end{figure}

\begin{algorithm}[t]
\caption{Closed-Loop Contextual Uncertainty Resolution}
\label{alg:framework}
\begin{algorithmic}[1]
\State \textbf{Input:} Task $\ell$, action budget $N$

\State $\mathcal{M}_0 \gets \emptyset, i \gets 1$, $o_0 \gets \emptyset, h_0 \gets \emptyset$

\While{$i \leq N$}

    \State $h_i, a_{i} \gets \pi_{LM}(o_{i-1}, h_{i-1}, \ell)$ 

    \If{$a_i$ \text{is} \texttt{hypothesis\_grounding}} 
        \State $(\mathcal{C}_i, \hat{a}_{1:C}) \gets \mathcal{G}(h_i, \mathcal{M}_i)$
        \State $o_i \gets \texttt{ToObservation}(\mathcal{C}_i, \hat{a}_{1:C})$

    \EndIf

    \If{$a_i$ \text{is} \texttt{active\_information\_gathering} }
        \State $o_i \gets \texttt{execute}(a_i)$ 
    \EndIf

    \State \textbf{If} $a_i = \texttt{answer(r)}$; \textbf{break}

    \State $i \gets i +1$
\EndWhile

\end{algorithmic}
\end{algorithm}

We address the above problem via \texttt{CLUE}, a closed-loop framework that pairs an LLM-enabled policy and a language-embedded mapping framework, as shown in Figure~\ref{fig:method}.
The policy has access to two categories of actions: \emph{hypothesis grounding} and \emph{active information gathering}.
In the first, the policy grounds hypotheses, which include task-relevant entities, into concrete action objectives.
In the second, the policy actively validates hypotheses via inspection, navigation, and manipulation, or otherwise makes progress towards the task.
The remainder of this section describes \texttt{CLUE}'s environment representations and observation space (\S\ref{sec:observation}), policy inference strategy (\S\ref{sec:planner}), hypothesis grounding method (\S\ref{sec:info_extraction}) and active information gathering (\S\ref{sec:env_interaction}).

\subsection{Environment Representations}
\label{sec:observation}
The primary challenge in designing an observation space is that the robot operates in a continuous, three dimensional world, while the LLM reasons over a symbolic (textual) space. 
\texttt{CLUE} addresses this by maintaining two levels of abstraction, each with distinct roles.
The \emph{language-embedded map} $\mathcal{M}_i = \{(x_j, y_j, z_j, f_j)\}$ is a dense voxel map whose feature embeddings $f_j$ are accumulated online from RGB-D imagery and may be queried for semantic relevance (\S\ref{sec:info_extraction}), and is augmented with updates $\Delta \mathcal{M}_i$ at each planning iteration.
It serves as the robot's perceptual memory, but it is inefficient to expose to the policy directly.
We thus define an observation graph, a sparse, symbolic abstraction comprising discrete entities and traversable regions annotated with coordinates and natural-language attributes.
The graph comprises two node types: \verb|object| and \verb|region|.
Objects denote semantic entities, while regions are points in freespace traversable by the robot.
Each node carries a coordinate and may be enriched with attributes in natural language, such as the output of a VLM query.
These updates, along with associated metadata, constitute the observation $o_i$, which is provided to the LLM via in-context updates from a textual API which denotes graph operations such as \verb|add_node()| or \verb|remove_edge()|.
For example, the second barrel from the example in Figure~\ref{fig:intro_figure} would yield the update:
\begin{tcolorbox}[
  colback=gray!3,
  colframe=black,
  left=1mm, right=1.5mm, top=1.5mm, bottom=1mm,
  minted language=text,
  minted options={breaklines, breakanywhere, fontsize=\scriptsize}
]
\texttt{add\_node(name=barrel\_2, coordinates=[5, -2], description=...)}

\texttt{add\_node(name=near\_barrel\_2, coordinates=[0,0])}

\texttt{add\_edge(barrel\_2, near\_barrel\_2)}
\end{tcolorbox}

\noindent The description of these objects would change throughout the task. 
For example, \verb|barrel_2| may start out with a description ``candidate location from language-embedded map,'' change to ``occluded by tarp,'' then finish with ``sign indicates barrel contains water.''

\subsection{Contextual Planning}
\label{sec:planner}
We instantiate our policy defined in \S\ref{sec:problem} with a pre-trained LLM that is configured with a system prompt that defines its role in \texttt{CLUE}.

\shortskip
\noindent \textbf{LLM Configuration.}
The LLM's system prompt specifies several components. 
It first describes the observation space and the action space (\ie behaviors and instructions for instantiating them with environmental entities).
It then provides a ``planning principles'' section which describes the sequential and closed-loop nature of the problem.
Finally, it provides instructions for constructing a hypothesis state and the required output format.

\shortskip
\noindent 
\textbf{Hypothesis state.}
At each planning iteration, the LLM emits a structured \emph{hypothesis state} $h_i$: a running, symbolic account of the task.
This hypothesis state tracks what the robot has confirmed (observed objects and regions), what it has hypothesized but not yet verified (inferred or candidate entities, together with hypotheses already ruled out), and the open uncertainties that remain task-relevant.
Carrying this state in the output lets the policy reason across iterations: confirmed entities ground later decisions, ruled-out hypotheses prevent repeated effort, and outstanding uncertainties drive the next round of information gathering.
A condensed schema is shown below.
\begin{tcolorbox}[
  colback=gray!3,
  colframe=black,
  left=1mm, right=1.5mm, top=1.5mm, bottom=1mm,
  minted language=text,
  minted options={breaklines, breakanywhere, fontsize=\small}
]
\{ "goal": "...",
   "confirmed": \{ "objects": [...], "regions": [...] \},
   "hypothesized": \{ "entities": [...], "ruled\_out": [...] \},
   "uncertainties": [...]
   \}
\end{tcolorbox}

\subsection{Hypothesis Grounding}
\label{sec:info_extraction}
\begin{figure}[t]
    \centering
    \includegraphics[width=1.0\linewidth]{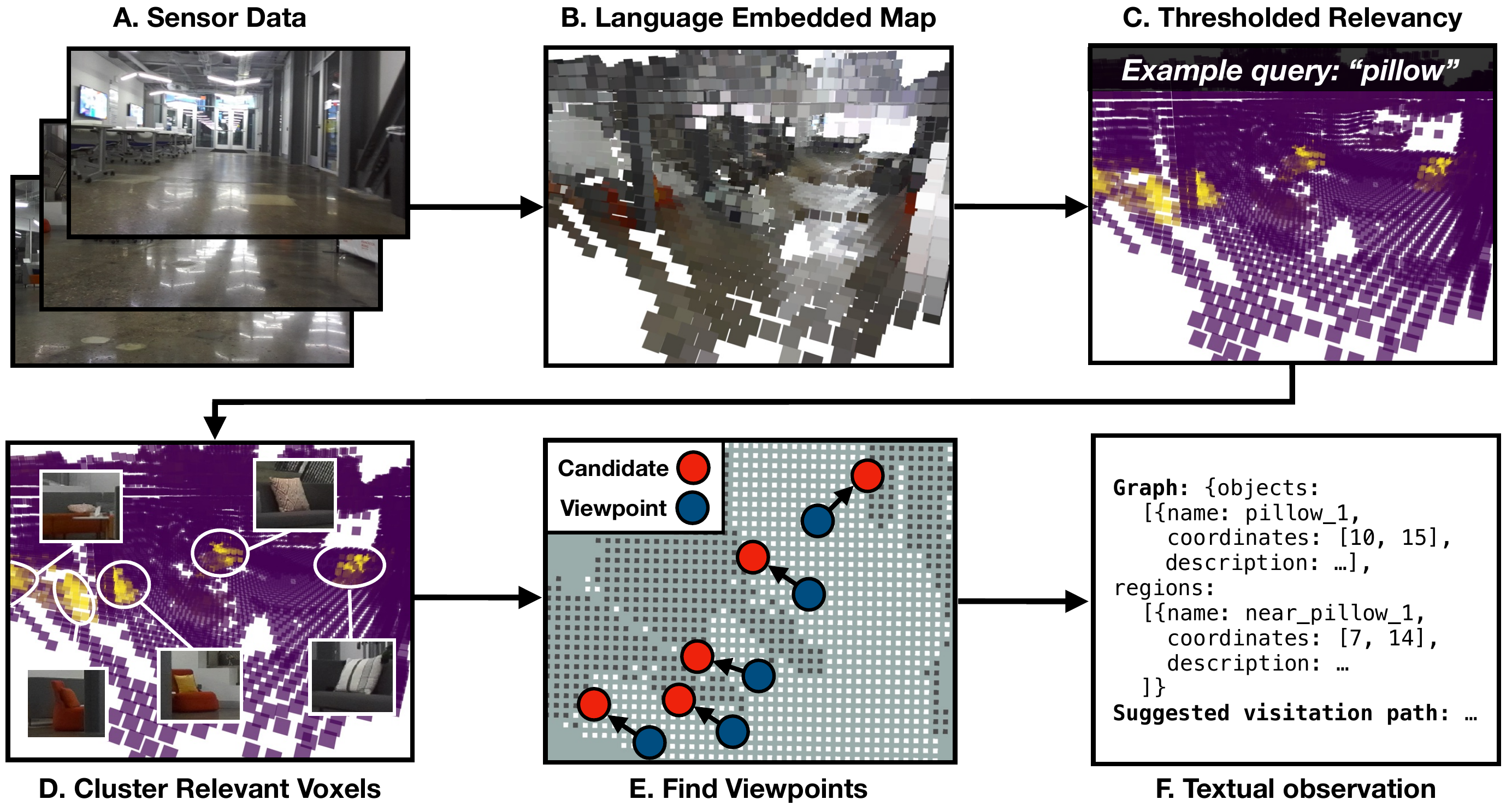}
    \caption{Hypothesis grounding. Sensory data is used to construct a language-embedded map. Given a concept from the policy's hypothesis, candidate entities are identified by finding high-relevancy voxels and clustering. Viewpoints from which the candidates may be observed are identified via the map's occupancy grid. These candidates and viewpoints form a graph which is provided to the policy via a textual observation along with a minimum distance visitation order.}
    \label{fig:info_extraction}
    \vspace{-12pt}
\end{figure}

The policy translates hypotheses into concrete planning objectives by invoking a grounding  process $\mathcal{G}(h_i, \mathcal{M}_i) \rightarrow \{\mathcal{C}_i, \hat{a}_{1:C}\}$.
As illustrated in Figure~\ref{fig:info_extraction}, this process takes as input a hypothesis and current language-embedded map $\mathcal{M}_i$, and it provides $C$ candidate inspection locations $\mathcal{C}_i$ along with a suggested distance-optimal visitation sequence $\hat{a}_{1:C}$.

\shortskip
\noindent\textbf{Map Querying.}
The policy triggers a grounding by identifying a task-relevant concept from a hypothesis and calling a \verb|query| action. 
This concept is then queried in the map, which is constructed online from sensor data (Fig.~\ref{fig:info_extraction}.A-B), by first encoding the concept into a feature vector, $f_c = \phi(c)$, then computing the cosine similarity between the concept and every feature vector in the map, $s_j = f_c^\top f_j \big/  {\lVert f_c\rVert\,\lVert f_j\rVert}, \; \forall f_j \in \mathcal{M}_i$.
Because this score is difficult to interpret as-is, we normalize it with a softmax against a set of background concepts (e.g., ``object,'' ``ground,'' ``wall''), yielding a confidence $\kappa_j =  e^{s_j} \big/  (e^{s_j} + \sum_{b} e^{s_{b,j}})$, where $s_{b,j}$ is the similarity of voxel $j$ to background concept $b$, and we retain all voxels whose confidence exceeds a threshold (Fig.~\ref{fig:info_extraction}.C).
We then cluster retained voxels using the DBSCAN algorithm~\cite{ester1996dbscan}, yielding a set of clusters $\{(x_m, y_m, n_m, \kappa_m)\}_{m=0}^M$, each comprising a centroid $(x_m, y_m)$, a size in voxels $n_m$, and an average confidence $\kappa_m$ (Fig.~\ref{fig:info_extraction}.D). 
The map's feature embeddings may not be precise enough for the task at hand.
For example, the task may require finding a ``red backpack,'' but only ``bag'' is identifiable by the embeddings.
The policy may therefore broaden its query if initial results come up empty.

\shortskip
\noindent
\textbf{Candidate selection and viewpoint planning.} 
\texttt{CLUE} only considers candidates with a confidence above threshold $\lambda$, yielding a set of candidates $\mathcal{C}_i=$\\
 $\{(x_c, y_c, n_c, \kappa_c)\}_{c=0}^C$. 
For each candidate, \textsc{CLUE} computes a viewpoint from which the robot can further inspect the candidate, if desired.
We define a target viewing distance---\ie how far the robot should stand from the object---and place a candidate viewpoint on the straight line between the robot's current pose and the target, at that distance from the object.
Finally, \texttt{CLUE} snaps this viewpoint to the nearest unoccupied point in freespace (Fig.~\ref{fig:info_extraction}.E).
Upon a query, the candidate entities and their viewpoints are added to the observation as \verb|object| and \verb|region| nodes (the process \texttt{ToObservation($\mathcal{C}_i, \hat{a}_{1:C})$} in Algorithm~\ref{alg:framework}).
\texttt{CLUE} also provides a suggested minimum-distance inspection tour to the policy via observation metadata, where $d_j$ is the distance incurred by the $j$-th action in the tour:
\begin{align}
\hat{a}_{1:C} = \operatorname*{arg\,min}_{a_{1:C}} \sum_{j=1}^{C} d_j.
\end{align}
\texttt{CLUE} computes a distance-optimal visitation sequence by solving a Traveling Salesman Problem (TSP), where distance is Euclidean distance between points, and it provides this ordering to the policy (Fig.~\ref{fig:info_extraction}.F).
The value of this hint is evaluated in \S\ref{sec:path_planning}.
However, this sequence may be missing necessary preconditions---for example the robot must navigate near a candidate before inspection or remove occlusion---it thus serves as a hint from which the policy can construct an action.

\subsection{Active Information Gathering}
\label{sec:env_interaction}

Having either queried the language-embedded map for candidates or acquired observations through direct environment interaction, the policy constructs and executes an action sequence to make task progress.
To realize the candidates as actions, the LLM infers the necessary preconditions, such as navigating to the associated viewpoint or clearing occlusion in order to properly inspect a candidate, as in Figure~\ref{fig:intro_figure}.
The policy replans as new information is discovered, and may terminate early once the task is resolved.

The policy constructs actions by instantiating behaviors with concrete goals in the environment. 
We consider six behaviors: \verb|goto()|, \verb|inspect()|, \verb|grasp()|, \verb|place()|, \verb|query()|, and \verb|answer()|.
goto() drives the robot to a location; inspect() calls a VLM with an LLM-generated query (e.g., ``does the barrel contain hazardous content'') and retrieves the response; grasp() and place() perform manipulation; and answer() terminates the task with a response to the user.
The policy instantiates actions using entities currently in the observation graph.
For example, the valid actions from the episode illustrated in Figure~\ref{fig:intro_figure} include \verb|goto(near_barrel_1)| and \verb|inspect(barrel_1)|.

\section{Experimental Evaluation}
\label{sec:experiment}
We design experiments to evaluate the contributions claimed in \S\ref{sec:intro}. Concretely, we aim to answer the following questions: 
\textbf{Q1.} Can \texttt{CLUE} realize tasks that require resolving contextual uncertainty?
\textbf{Q2.} Is closed-loop feedback required, or does open-loop planning suffice? \textbf{Q3.} Is \texttt{CLUE}'s grounding process necessary, or does simply building and then querying a semantic map suffice? 
We describe our implementation details in \S\ref{sec:implementation}, our experimental setup---including environments, tasks, and baselines---in \S\ref{sec:setup}.
We report results in \S\ref{sec:main_results}, and we provide supporting experiments in 
\S\ref{sec:exp_info_retreival} and \S\ref{sec:path_planning}.
\subsection{Implementation details}
\label{sec:implementation}

We implement our mapping framework by extending RayFronts~\cite{alama2025rayfronts} with the RadSeg feature encoder~\cite{alama2025radseg}.
We use SciPy's implementation of DBSCAN~\cite{2020SciPy-NMeth} and implement a TSP solver using Gurobi~\cite{gurobi}.
We implement our policy and its VLM using GPT-5.1 with ``low'' reasoning~\cite{openai2025gpt5systemcard}. 
We run all experiments on the Boston Dynamics Spot, which is equipped with a forward-facing ZED 2i camera for RGB and depth sensing, an NVIDIA Jetson AGX Thor for compute, and provides odometry via internal state estimation.
Mobility relies on Nav2 for trajectory planning and the Spot's internal controller for velocity tracking~\cite{macenski2020marathon2}.
Similarly, we use the Spot's internal SDK for object grasping and placement.
We query GPT-5.1 via a network connection; all other processes run onboard.

\begin{figure}[t]
    \centering
    \includegraphics[width=1.0\linewidth]{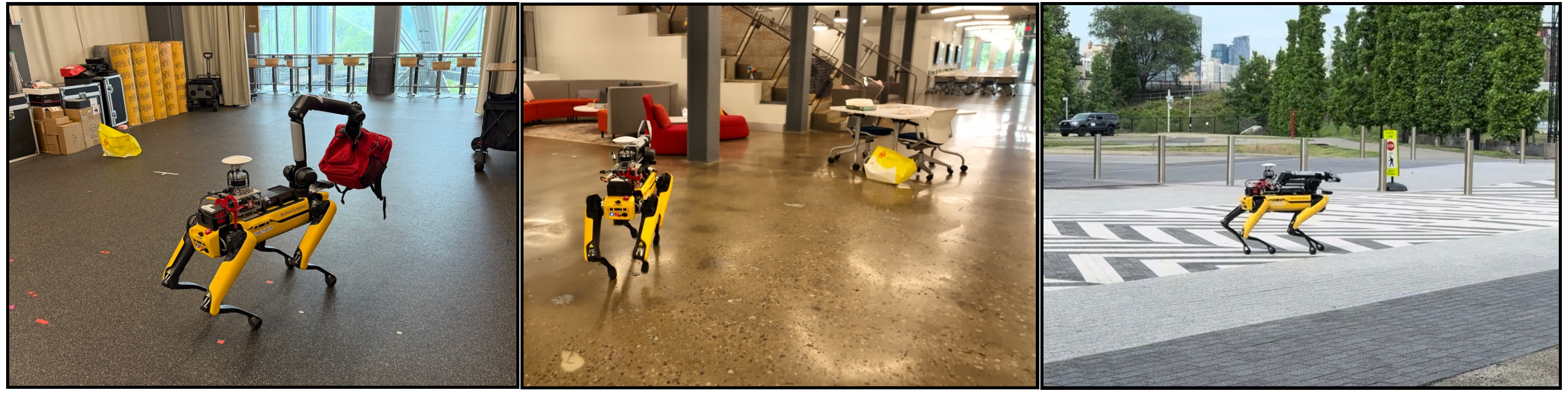}
    \caption{Experimental snapshots showing the evaluation environments: two indoor settings and one outdoor setting.}
    \label{fig:environments}
\end{figure}

\subsection{Experimental setup}
\label{sec:setup}

This section describes the experimental environments considered in our work, tasks, and baselines.

\shortskip
\noindent
\textbf{Environments.} We conduct experiments in two indoor environments and the outside of an office park, which are illustrated in Figure~\ref{fig:environments}. 
These environments provide an experimental area of roughly 100m$^2$, 250m$^2$, and 1,000m$^2$.

\shortskip
\noindent
\textbf{Tasks.}
We consider three types of tasks: object disambiguation, functional inference, and occlusion reasoning. 
Object disambiguation tasks require \texttt{CLUE} to identify a specific description of an object (e.g., a Jansport bag). 
The feature embeddings of the language-embedded map do not provide a sufficient level of detail to locate the object, and the map may have false positives. \texttt{CLUE} must therefore use the map to gather candidate inspection locations and then verify candidates online.
\texttt{CLUE} must then situate this object within the context of the task (\eg mobile manipulation).
Functional inference tasks suggest desired object attributes, such as ``I need something to transport equipment for field experiments. Ideally with a perch for my laptop.'' 
\texttt{CLUE} must infer entities that match the functional attributes, \eg a cart for transport and some supporting structure for a laptop.
Finally, occlusion reasoning requires \texttt{CLUE} to recognize when an object of interest may be occluded and resolve that occlusion, as illustrated in Figure~\ref{fig:intro_figure}.
These tasks require both manipulation---for occlusion reasoning and acquiring objects of interest---and mobility. 
We consider a total of fifteen tasks---nine in the laboratory environment, three in a floor of the office building, and three outdoors---and we split these tasks into six object disambiguation, six functional inference, and three occlusion reasoning. We evaluate each task once.

\shortskip
\noindent
\textbf{Baselines.}
We consider two baselines. 
First, we evaluate against an oracle which is given step-by-step instructions on how to solve the task, but otherwise has the same action space as our method.
This baseline estimates an upper bound on task performance.
We also compare against NLMaps~\cite{chen2023open}. 
This policy has access to the same language-embedded map and action space as our method, but it  assumes queries from the map are noiseless.
The policy must therefore generate an open-loop plan, and this baseline measures the importance of closed-loop feedback for contextual information gathering.

\begin{table}[t]
    \centering
    \scriptsize
    \begin{tabular}{c|ccc|cccccc}
        \toprule
        \multirow{2}{*}{Method}& \multicolumn{3}{c|}{Success} & \multirow{2}{*}{Actions} & \multirow{2}{*}{Time (s)} & \multirow{2}{*}{LLM Calls} & \multirow{2}{*}{LLM Tok. (k)} & \multirow{2}{*}{VLM Calls} & \multirow{2}{*}{VLM Tok. (k)} \\
         \cmidrule(lr){2-4} &  Dis.  & Func. & Occl. \\ \toprule 
         Oracle & 83.3\% & 100.0\% & 100.0\% & 4.3$\pm$1.0 & 175.4$\pm$62.0 & 4.4$\pm$1.2 & 2.3$\pm$8.0 & 1.1$\pm$0.5 & 0.9$\pm$0.4 \\ 
         \texttt{CLUE} & 83.3\% & 100.0\% & 66.7\% & 8.1$\pm$4.9 & 222.9$\pm$120.6 & 6.3$\pm$3.6 & 5.7$\pm$5.2 & 2.9$\pm$2.1 & 2.1$\pm$1.7\\ 
         NLMaps~\cite{chen2023open} & 16.7\% & 33.3\% & 0.0\% & 4.3$\pm$1.2 & 115.2$\pm$58.4 & 3.4$\pm$0.8 & 1.2$\pm$4.6 & 0.9$\pm$0.6 & 0.5$\pm$0.3 \\
         \bottomrule 
    \end{tabular}
    \vspace{3pt}
    \caption{Baseline comparison mean and std. across fifteen real-world experiments, with success broken down by task type: object disambiguation (Dis.), functional inference (Func.), and occlusion reasoning (Occl.).}
    \label{tab:main}
\end{table}

\subsection{Results}
\label{sec:main_results}

Table~\ref{tab:main} reports results across seven metrics: task success rate, the number of actions required to finish a task, the time required to terminate a task, the number of LLM calls, the number of LLM tokens (LLM Tok.), the number of VLM calls, and the number of VLM tokens (VLM Tok.) used.
The oracle achieved a 93.3\% success rate. Failure occurred when a candidate was mislocalized in the map due to depth reconstruction error, and the policy could not recover despite additional exploration attempts.
\texttt{CLUE} achieved an 86.7\% success rate averaged over tasks, addressing \textbf{Q1.}
Resource usage---as measured by the remaining six metrics---is significantly higher. \texttt{CLUE} takes 1.3x as long with nearly 2x as many actions, 1.4x as many LLM calls with 2.5x as many LLM tokens, 2.6x as many VLM calls, and 2.3x as many VLM tokens. 
NLMaps only achieves a success rate of 20\%, addressing \textbf{Q2.} The resource usage is comparable or lower than the oracle's, with the same number of actions on average, 0.7x as long, 0.8x as many LLM calls with nearly half the number of tokens, and similar trends for VLM usage.

\begin{figure}[t]
    \includegraphics[width=0.95\textwidth]{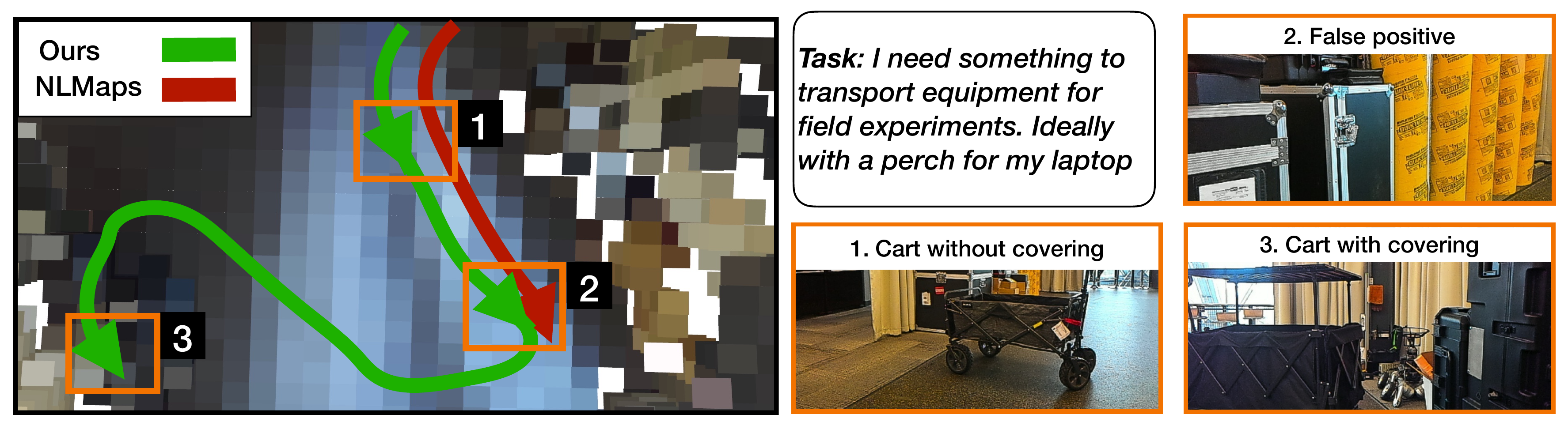}
    \caption{\textbf{Qualitative Results.} Our method sequentially inspects candidates derived from the map, reasoning about functional use until it arrives at a suitable object. NLMaps only visits one candidate, which is a false positive.}
    \label{fig:qual_results}
    \vspace{-12pt}
\end{figure}

\texttt{CLUE} uses more resources (actions, time, LLM/VLM calls and tokens) because it had to resolve false positives and distractors in the map. In contrast, the oracle received ground truth instructions so could directly address task-relevant content.
\texttt{CLUE} had two primary failure modes: contextual inference and behavior execution failure. 
Contextual inference failure came when the policy incorrectly interpreted the task. For example, in response to the task ``I need something to carry my robot outside,'' it suggested a packing box rather than a cart. 
In another experiment, the robot failed to grasp a task-relevant object, leading to an execution failure.
NLMaps' primary failure mode was false positives in the initial map query. 
It assumed this result was correct, so it had no mechanism to recover from false positives (see Figure~\ref{fig:qual_results} for an example).
Because NLMaps would simply stop after reaching a false positive, rather than trying to explore additional locations, its resource usage is comparatively low.

\subsection{Does Information Retrieval on a Map Suffice?}
\label{sec:exp_info_retreival}

We next evaluate the necessity of active perception as opposed to information retrieval from a pre-built map. 
We compare \texttt{CLUE} to DAAAM~\cite{Gorlo2025DAAAM}, a state-of-the-art scene understanding framework which differs from our approach in two key aspects.
First, DAAAM forgoes a dense language-embedded map and instead directly constructs a scene graph---explicitly segmenting objects and regions.
Second, DAAAM uses a VLM to caption objects and regions observed during map construction. Because DAAAM does not receive a prior task, these captions are descriptive but generic.
We compare \texttt{CLUE} to DAAAM across the nine tasks performed in the laboratory environment.
We first use the sensory data collected during the experiments from \S\ref{sec:main_results} to build a map of the environment.
We then provide DAAAM's agentic information retrieval tool with the task specification.

Table~\ref{tab:map_experiment} reports the mean and standard deviation over five metrics---task success, LLM calls, LLM tokens used, VLM calls, and VLM tokens used.
We find that DAAAM only achieves a 22.2\% task success rate averaged across tasks, addressing \textbf{Q3.} The primary failure modes were contextual imprecision where the map did not have enough information to disambiguate objects, false negative object detections which prevented task-relevant object retrieval, and the inability to resolve occlusions in the scene.
Because \texttt{CLUE} uses an LLM to repeatedly generate new plans, it requires over twice as many LLM calls, with higher variance resulting from variable task length. 
Interestingly, \texttt{CLUE} requires only 1.2x as many tokens despite a higher number of queries; this is because DAAAM's agent requires the entire scene graph, which is significantly larger than the symbolic observation space described in \S\ref{sec:observation}.
Finally, because DAAAM is using a VLM to caption all observed objects rather than only ones that are contextually-relevant, it requires over 2.8x as many VLM calls with 12x as many tokens.

\begin{table}[t]
    \centering
    \begin{tabular}{c|ccc|cccc}
        \toprule
        \multirow{2}{*}{Method}& \multicolumn{3}{c|}{Success} & \multirow{2}{*}{LLM Calls} & \multirow{2}{*}{LLM Tok. (k)} & \multirow{2}{*}{VLM Calls} & \multirow{2}{*}{VLM Tok. (k)} \\
         \cmidrule(lr){2-4} & Dis. & Func. & Occl. \\ \toprule
         \texttt{CLUE} & 66.7\% & 100.0\% & 66.7\% & 5.2$\pm$3.1 & 4.2$\pm$4.4 & 1.9$\pm$1.3 & 1.4$\pm$1.2 \\
         DAAAM~\cite{Gorlo2025DAAAM} & 33.3\% & 33.3\% & 0.0\% & 2.1$\pm$0.6 & 3.4$\pm$0.7 & 5.4$\pm$0.5 & 17.1$\pm$2.4\\
         \bottomrule 
    \end{tabular}
    \vspace{3pt}
    \caption{Map retrieval mean and std. across nine real-world experiments, with success broken down by task type: object disambiguation (Dis.), functional inference (Func.), and occlusion reasoning (Occl.).}
    \label{tab:map_experiment}
    \vspace{-24pt}
\end{table}

\subsection{Can LLMs Provide Informative Paths?}
\label{sec:path_planning}

We next assess the value of providing the LLM with a plan generated by a TSP (\S\ref{sec:info_extraction}) by comparing \texttt{CLUE} with an ablated version that does not receive plan hints (\texttt{CLUE} w/o hint), an LLM that has been prompted to produce a distance-efficient path (Path LLM), and the result from TSP.
Each method must plan a path over waypoints randomly sampled from a 10$\times$10 grid, which corresponds to candidates from \texttt{CLUE}'s hypothesis grounding process.
We evaluate the ability of each method to produce distance-optimal paths from two to ten waypoints.
For ``\texttt{CLUE}'' and ``\texttt{CLUE} w/o hint,'' we append the generated points to the observation history from a recorded experiment to mimic deployment.
\begin{wrapfigure}{r}{0.5\textwidth}
    \vspace{-6pt}
    \centering
    \includegraphics[width=1\linewidth]{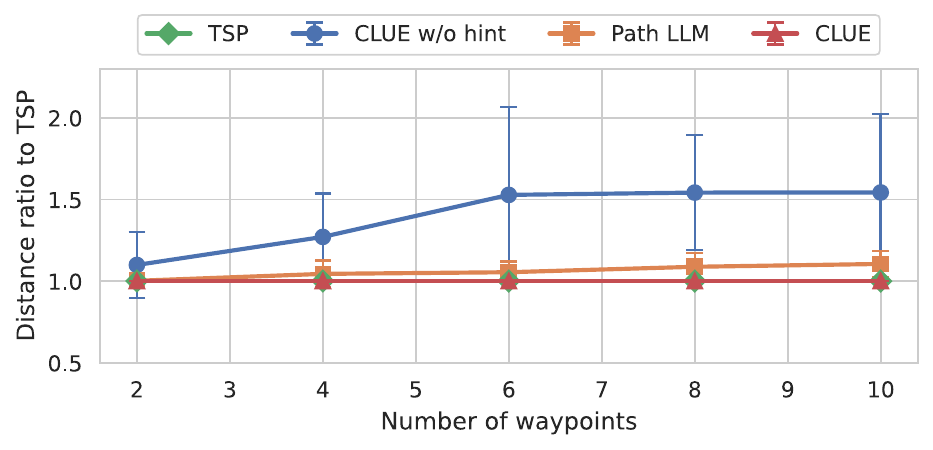}
    \caption{LLM ability to plan distance-efficient paths (mean and std.)}
    \label{fig:planning_performance}
    \vspace{-6pt}
\end{wrapfigure}
Figure~\ref{fig:planning_performance} reports the mean and standard deviation of the distance ratio to the TSP solution averaged over 25 trials from two to ten waypoints.

The Path LLM is able to rival the performance of the TSP, which is a result suggested by~\cite{huang2024canllmpath}.
Interestingly, without a TSP hint our method's LLM produces significantly longer paths---over 1.5 times as long for six or more points. 
This degradation is likely due to the additional system instructions---as outlined in \S\ref{sec:planner}---which focus the LLM's attention on making contextual associations in the environment rather than producing distance-efficient paths.
Future work is required to develop generation structures that enable both reasoning and efficient path assignments.

\section{Limitations and Future Work}
\label{sec:limitations}
We note several limitations and areas for future work.
The information provided in the language-embedded map may be used for semantic frontier exploration, such as in \cite{ong2025atlas}, and this is a promising avenue for future work.
The language-embedded map is also memory intensive, with maps for the outdoor environment reaching 100 GB. 
Existing work suggests that memory requirements may be reduced via submap caching, which stores inactive portions of the map to disk~\cite{ong2025atlas}.

Finally, we only consider a cloud-based model (\ie GPT-5.1). 
Future work may evaluate the performance of \texttt{CLUE} using smaller open-source models or consider distillation methods in order to produce on-device policies~\cite{ravichandran_prism}.

\section{Conclusion}
\label{sec:conclusion}

This work addresses the problem of resolving contextual uncertainty for underspecified tasks in natural language, where the required subgoals as well as the relevant semantic, functional, and metric information must be inferred.
We propose \texttt{CLUE}, a closed-loop framework that uses an LLM-enabled policy to infer hypotheses, an approximation of belief for open-vocabulary environments, given a task description and observations.
\texttt{CLUE} grounds these hypotheses into concrete actions using a language-embedded map, and verifies hypotheses through active interaction with the environment.
These interactions produce observations, which the policy uses to update its hypotheses and iteratively replan. 
Across real-world experiments in indoor and outdoor environments that require object disambiguation, functional affordance reasoning, and occlusion handling, \texttt{CLUE} rivals the performance of an oracle with ground truth instructions (within 7 percentage points task success), outperforms an LLM-enabled planner without closed-loop information gathering by a 4x margin and a mapping then querying approach by a 3x margin, while being significantly more VLM token efficient.

\vspace{6pt}
\noindent \textbf{Acknowledgments.} We gratefully acknowledge support from DCIST CRA \\W911NF-17-2-0181, NSF Grant CCF-2112665, and the NSF GRFP.

% \printbibliography

\bibliographystyle{spmpsci}
\bibliography{refs}

\end{document}